\documentclass[11pt]{article}
\usepackage{framed}
\usepackage{amssymb,amsfonts,amsmath}
\usepackage{algorithm2e,framed}
\usepackage{amsfonts,amsmath}
\usepackage{graphicx}
\usepackage{subfigure,enumerate,bm,amsmath,amsthm,wrapfig,array,
color,algorithmic}
\usepackage{graphics}
\usepackage{multirow}
\usepackage{setspace}

\long\def\symbolfootnote[#1]#2{\begingroup%
\def\thefootnote{\fnsymbol{footnote}}\footnote[#1]{#2}\endgroup}

\newcommand{\Expect}[1]{\mbox{}{\mathbb{E}}\left[#1\right]}

\newcommand{\FNorm }[1]{\mbox{}\|#1\|_\mathrm{F}  }
\newcommand{\FNormS}[1]{\mbox{}\|#1\|_\mathrm{F}^2}

\newcommand{\TNorm }[1]{\mbox{}\|#1\|_2  }
\newcommand{\TNormS}[1]{\mbox{}\|#1\|_2^2}

\newcommand{\XNorm }[1]{\mbox{}\|#1\|_{\xi}  }

\newtheorem{theorem}{\bf Theorem}[]
\newtheorem{lemma}[theorem]{Lemma}
\newtheorem{definition}[theorem]{Definition}

\newtheorem{corollary}[theorem]{Corollary}

\newcommand{\transp}{^{\textsc{T}}}

\newcommand{\mat}[1]{{\ensuremath{\bm{\mathrm{#1}}}}}

\def\rank{\hbox{\rm rank}}

\def\b{{\mathbf b}}

\def\p{{\mathbf p}}
\def\q{{\mathbf q}}

\def\v{{\mathbf v}}

\def\matA{\mat{A}}
\def\matB{\mat{B}}
\def\matC{\mat{C}}

\def\matE{\mat{E}}

\def\matI{\mat{I}}

\def\matQ{\mat{Q}}

\def\matS{\mat{S}}

\def\matU{\mat{U}}
\def\matV{\mat{V}}

\def\matX{\mat{X}}
\def\matY{\mat{Y}}
\def\matZ{\mat{Z}}
\def\matSig{\mat{\Sigma}}

\def\matOmega{\mat{\Omega}}
\def\matPsi{\mat{\Psi}}

\def\scl{{\textsc{l}}}
\def\scu{{\textsc{u}}}

\DeclareMathSymbol{\Prob}{\mathbin}{AMSb}{"50}
\newcommand\remove[1]{}

\def\math#1{$#1$}

\def\mand#1{$$#1$$}
\def\frac#1#2{{#1\over #2}}

\def\mld#1{\begin{equation}
#1
\end{equation}}
\def\eqar#1{\begin{eqnarray}
#1
\end{eqnarray}}
\def\eqan#1{\begin{eqnarray*}
#1
\end{eqnarray*}}

\DeclareMathSymbol{\R}{\mathbin}{AMSb}{"52}

\def\cl#1{{\cal #1}}

\newcommand{\argmin}{\operatorname*{argmin}}

\def\z{{\mathbf z}}

\def\b{{\mathbf b}}

\def\norm#1{{\|#1\|}}

\def\r#1{{(\ref{#1})}}

\def\dotfil{\leaders\hbox to 1.5mm{.}\hfill}
\newcounter{rmnum}
\def\RN#1{\setcounter{rmnum}{#1}\uppercase\expandafter{\romannumeral\value{rmnum}}}
\def\rn#1{\setcounter{rmnum}{#1}\expandafter{\romannumeral\value{rmnum}}}

\begin{document}

\title{\bf Deterministic Feature Selection for K-means Clustering}

\author{
{\bf Christos Boutsidis} \\
Mathematical Sciences Department \\
IBM T.J. Watson Research Center \\
cboutsi@us.ibm.com
\and
{\bf Malik Magdon-Ismail}\\
Computer Science Department \\
Rensselaer Polytechnic Institute \\
magdon@cs.rpi.edu
}
\maketitle

\begin{spacing}{1.25}

\begin{abstract}
We study feature selection for $k$-means clustering.~Although the literature contains many methods with good empirical performance,
algorithms with provable theoretical behavior have only recently been developed.~Unfortunately, these algorithms are randomized and fail with, say, a constant probability.
We present the first \emph{deterministic} feature selection algorithm
for $k$-means clustering with relative error guarantees.
At the heart of our algorithm lies a deterministic method
for decompositions of the identity and a structural result which quantifies
some of the tradeoffs in dimensionality reduction.
\end{abstract}

\section{Introduction}
This paper is about feature selection for $k$-means clustering, a
topic that received considerable attention from scientists and engineers.
Arguably, $k$-means is the most widely used clustering algorithm in practice~\cite{Wu07}.
Its simplicity and effectiveness are remarkable among all the available
methods~\cite{ORSS06}. On the negative side, using $k$-means to cluster high dimensional data
with, for example, billions of features is not simple and straightforward~\cite{GGBD05};
the curse of dimensionality makes the algorithm very slow.
On top of that, noisy features often lead to overfitting, another undesirable effect.
Therefore, reducing the dimensionality of the data by selecting a subset of the features, i.e. feature selection,
and optimizing the $k$-means objective on the low dimensional representation of the high dimensional
data is an attractive approach that not only will make $k$-means faster, but also more robust~\cite{GE03,GGBD05}.

The natural concern with throwing away potentially useful dimensions is that
it could lead to a significantly higher clustering error. So, one has to select the features
carefully to ensure that one can recover comparably good clusters just by
using the dimension-reduced data.  Practitioners have developed numerous
feature selection methods that work well \emph{empirically}~\cite{GE03,GGBD05}.
The main focus of this work is on algorithms for feature selection with
provable guarantees.
Recently, Boutsidis
et al.
described a feature selection algorithm that gives a theoretical
guarantee on the quality of the clusters that are produced after
reducing the dimension~\cite{BMD09c}. Their algorithm, which
employs a technique
of Rudelson and Vershynin~\cite{RV07}, selects the features
randomly with probabilities that
are computed via the right singular vectors of the matrix containing the data
(\cite{BZMD11} describes a similar randomized algorithm with the same bound but faster running time). Although Boutsidis et al. give a strong theoretical bound for the quality of the resulting clusters (we will discuss this bound in detail later),
the bound fails with some non negligible probability, due to the randomness in how they sample the features.
This means that every time the feature selection is performed, the algorithm could (a) fail, and
(b) return a different answer each time.
To better address the applicability of such feature selection algorithms for $k$-means, there is a need for \emph{deterministic}, provably accurate feature selection algorithms. We present the first deterministic algorithms of this type.
Our contributions can roughly be summarized as follows.
\begin{itemize}
\item{\bf Deterministic Supervised Feature Selection (Theorem~\ref{thm1}).}
Given \emph{any} dataset and \emph{any} input
$k$-partition of this dataset, there is a small set of $O(k)$
feature dimensions in which any near optimal output
\math{k}-partition of the data
 is no more than a constant factor worse in quality than the given input
 $k$-partition (the quality of the input and output \math{k}-partitions
are compared in the original dimension).
Moreover, this small set of feature dimensions can be computed in deterministic low-order polynomial time. This is the first deterministic algorithm of this type. Prior work~\cite{BMD09c,BZMD11} gives a randomized algorithm that can
only reduce to
$\Omega(k \log k)$ dimensions and guarantee comparable clustering quality.
\item{\bf Existence of a small set of near optimal features (Corollary~\ref{col}).}
We prove existence of a small set of near optimal features. That is, given \emph{any} dataset and the number of clusters $k$,
there is a set of $O(k)$ feature dimensions
such that any optimal $k$-partition in the reduced dimension
is no more than a constant factor
worse in quality than the optimal $k$-partition of the dataset. The existence of such a small set of features was not known before. Prior work~\cite{BMD09c,BZMD11}
only  implies the existence of $\Omega(k \log k)$ features with comparable performance.
\item{\bf Deterministic Unsupervised Feature Selection (Theorem~\ref{thm2}).}
Given \emph{any} dataset and the number of clusters $k$, it is possible, in deterministic low-order polynomial time, to select $r$ feature dimensions, for any $r>k$, such that the optimal $k$-partition of the dimension-reduced data is no more than $O(n/r)$ worse in quality than the optimal $k$-partition; here $n$ is the number of features of the dataset. This is the first deterministic algorithm of this type. Prior work~\cite{BMD09c,BZMD11} offers a randomized algorithm with  error
$\left( 3 + O(k \log k / r) \right)$, for $r = \Omega(k \log k)$.
\item{\bf Unsupervised Feature Selection with a small subset of features (Theorem~\ref{thm3}).}
Finally, given \emph{any} dataset and the number of clusters $k$,
in randomized low-order polynomial time it is possible to select
a small number, $r$, of
 feature dimensions, with $k < r = o(k \log k)$, such that the
optimal $k$-partition for the dimension-reduced data is
no more than $O(k\log (k)/r)$ worse in quality than the optimal $k$-partition.
In particular,
this is the first (albeit randomized) algorithm of this
 type that can select a small subset of $O(k)$ feature dimensions
and provide an \math{O(\log k)}-factor guarantee.
Prior work~\cite{BMD09c,BZMD11} is limited to selecting
$r = \Omega(k \log k)$ features. The new algorithm combines ideas from this paper with the technique of~\cite{BMD09c,BZMD11}.
\end{itemize}
In order to prove our results we prove a structural result in
Lemmas~\ref{lem:generic2} and~\ref{lem:generic}. This structural result
quantifies the tradeoffs when selecting feature dimensions in terms of how
the feature selection process preserves the matrix norms of certain
crucial matrices. This is a general structural result and may be of independent
interest.

In order to get deterministic algorithms and be able to select $O(k)$ feature dimensions, we need to use techniques that are
completely different from those used in~\cite{BMD09c,BZMD11}. Our approach is inspired by a recent deterministic result
for decompositions of the identity which was introduced in~\cite{BSS09} and subsequently extended in~\cite{BDM11a},
while~\cite{BMD09c,BZMD11} use the randomized technique of~\cite{RV07} to extract the features.
This general approach might also be of interest to the Machine Learning and Pattern Recognition communities,
with potential applications to other problems involving subsampling,
for example sparse PCA and matrix approximation~\cite{SS00,DGJL07}.

\subsection{Background}
We first provide the basic background on $k$-means clustering that is needed
to describe our results in Section~\ref{sec:main}.
We postpone the more technical background needed for
describing our main algorithms and for
 proving our main results to Section~\ref{sec:pre}.
We begin with
the definition of the $k$-means clustering problem.\footnote{In Section~\ref{sec:pre}, we provide an alternative definition using matrix notation, which will be
useful in proving the main results of this work.}
Consider a set \math{\cl{P}}
 of $m$ points in an $n$-dimensional Euclidian space,
$$\mathcal{P} = \{ \p_1, \p_2, ..., \p_m \} \in \R^{m\times n},$$
and an integer
$k$ denoting the desired number of clusters. The objective of
$k$-means is to find a $k$-partition of
$\mathcal{P}$ such that points that are ``close'' to each other belong to the same cluster and points
that are ``far'' from each other belong to different clusters.
A $k$-partition of $\mathcal{P}$ is a collection
$$\cl S=\{\mathcal{S}_1, \mathcal{S}_2, ..., \mathcal{S}_k\},$$
of \math{k} non-empty pairwise disjoint
sets which covers \math{\cl P}.
Let $s_j=|\mathcal{S}_j|,$ be the size of $\mathcal{S}_j$.
For each set \math{S_j}, let \math{\bm\mu_j\in\R^n} be its centroid
(the mean point),
$$\bm\mu_j=\frac{1}{s_j}\sum_{\p_i\in S_j}\p_i.$$
The $k$-means objective function is
$$
\mathcal{F}(\mathcal{P}, \cl S) =
\sum_{i=1}^m\norm{\p_i-\bm\mu(\p_i)}_2^2,
$$
where \math{\bm\mu(\p_i)} is the centroid of the cluster to which
\math{\p_i} belongs.
The goal of $k$-means is to find a partition
\math{\cl S} which minimizes \math{\cl F} for a given $\mathcal{P}$ and $k$. We will
refer to any such optimal clustering as,
$$\cl{S}_{opt} = \argmin_{\cl{S}} \cl{F} (\mathcal{P}, \cl{S}).$$
The corresponding objective value is $$\cl F_{opt} = \cl{F} (\mathcal{P}, \cl{S}_{opt}).$$
The goal of feature selection is to construct  points
$$\mathcal{\hat{P}} = \{ \hat{\p}_1, \hat{\p}_2, \dots, \hat{\p}_m \} \in \R^{m\times r},$$
(for some $r < n$ specified in advance)
by projecting each $\p_i$ onto $r$ of the coordinate dimensions.
Consider the optimum $k$-means partition of the points in $\mathcal{\hat{P}}$,
$$\hat{\cl{S}}_{opt} = \argmin_{\cl{S}} \cl{F} (\mathcal{\hat{P}}, \cl{S}).$$
The goal of feature selection is to construct a new set of points $\mathcal{\hat{P}}$ such that,
$$ \cl{F} (\mathcal{P}, \hat{\cl{S}}_{opt}) \leq \alpha \cdot \cl{F} (\mathcal{P}, \cl{S}_{opt}) .$$
Here, $\alpha$ is the approximation factor and might depend on $m,n,k$ and $r$.
In words, computing a partition $\hat{\cl{S}}_{opt}$ by using the low-dimensional data and plugging
it back into the clustering metric \math{\cl{F}(\cl{P},\cdot)} for
 the high dimensional data, gives an \math{\alpha}-approximation to the optimal value
of the $k$-means objective function.
Notice that we measure the quality of $\hat{\cl{S}}_{opt}$ by evaluating the $k$-means objective function in the original space,
an approach which is standard~\cite{OR99,KSS04,AV07}. Comparing $\hat{\cl{S}}_{opt}$ directly to
$\cl{S}_{opt}$, i.e. the identity of the clusters, not just the clustering error,
would be much more interesting but at the same time a much harder (combinatorial) problem.
A feature selection algorithm is called \emph{unsupervised} if it computes $\mathcal{\hat{P}}$ by only
looking at $\mathcal{P}$ and $k$. \emph{Supervised} algorithms construct $\mathcal{\hat{P}}$ with
respect to a given partition $\cl{S}_{in}$ of the data. Finally, an algorithm will be
a $\gamma$-approximation for $k$-means ($\gamma \ge 1$) if it finds a clustering $\cl S_{\gamma}$
with corresponding value $ \cl F_{\gamma} \le \gamma \cl F_{opt}.$
Such algorithms will be used to state our results in a more general way.

\begin{definition} \label{def:approx}
\textsc{[k-means approximation algorithm]}
An algorithm is a ``$\gamma$-approximation'' for $k$-means clustering ($\gamma \geq 1$) if
it takes as input the dataset $\mathcal{P} = \{ \p_1, \p_2, ..., \p_m \} \in \R^{m\times n}$ and the number of clusters $k$,
and returns a clustering \math{\cl S_{\gamma}} such that,
$$
\cl F_{\gamma} = \cl{F} (\mathcal{P}, \cl{S}_{\gamma}) \le \gamma \cl F_{opt}.
$$
\end{definition}
The simplest algorithm with $\gamma = 1$,
but exponential running time,
would try all possible $k$-partitions and return the best.
Another example of such an algorithm is in~\cite{KSS04}
with $\gamma = 1+\epsilon$ ($0 < \epsilon < 1$).
The corresponding running time is $O(m n \cdot 2^{(k/\epsilon)^{O(1)}})$.
Also, the work in~\cite{AV07} describes a method with
$\gamma = O(\log k)$
and running time $O(mnk)$. The latter two algorithms are randomized.
For other $\gamma$-approximation algorithms,
see~\cite{OR99} as well as the discussion and the references in~\cite{OR99,KSS04,AV07}.

\section{Statement of our main results} \label{sec:main}

We first discuss guarantees for supervised feature selection when the
user has a candidate input \math{k}-partition. We next present results
for the unsupervised case. We end this section with a brief discussion of the
results.

\subsection{Supervised Feature Selection}
Our first result is within the context of supervised feature selection.
Suppose that we are given points $\mathcal{P}$ and some $k$-partition \math{\cl S_{in}}.
The goal is to find the features of the points in $\mathcal{P}$ from which we can find a partition \math{\cl S_{out}}
that is not much worse than the given partition \math{\cl S_{in}}. Notice that if \math{\cl S_{in}} is arbitrarily bad,
then \math{\cl S_{out}} might be much better than \math{\cl S_{in}}; however, our bound does not capture how much better the
resulting partition would be. It only guarantees that it will not be much worse than the given partition.
\begin{theorem}
\label{thm1}
There is an $O(m n \min\{m,n\} + rk^2n)$ time deterministic
feature selection algorithm which takes as input
 any set of points
$\mathcal{P} = \{ \p_1, \p_2, ..., \p_m \} \in \R^{m\times n}$,
a number of clusters $k$, a number of features to be selected
$k < r < n$, and a $k$-partition of the points \math{\cl S_{in}}.
The algorithm constructs
$\mathcal{\hat{P}} = \{ \hat{\p}_1, \hat{\p}_2, ..., \hat{\p}_m \} \in \R^{m\times r}$ such that \math{\cl S_{out}},
an arbitrary $\gamma$-approximation
on $\mathcal{\hat{P}}$, satisfies
\eqan{
\cl F(\mathcal{P},\cl S_{out}) &\le&
\left(1+ \frac{4 \gamma}{(1 - \sqrt{k/r} )^{2}}\right)
\cl F(\mathcal{P},\cl S_{in}).\\
}
\end{theorem}
Asymptotically, as \math{r\rightarrow\infty}, the approximation factor is
\math{\gamma\cdot O\left(1+ \sqrt{k/r}\right)}.
Essentially the clustering $\cl S_{out}$ is at most a constant factor worse than
the original clustering $\cl S_{in}$; but, \math{\cl S_{out}} was computed using
the lower dimensional data. This means we can compress the number
 of the feature
dimensions without destroying a given specific clustering in the data.
This is useful, for
example, in privacy preserving applications where one seeks to release minimal information
of the data without destroying much of the encoded information~\cite{VC03}.
Notice that the feature selection part of the theorem (i.e. the construction of $\mathcal{\hat{P}}$)
is deterministic.
The $\gamma$-approximation algorithm, which can be randomized, is only used
to describe the  clustering that can be obtained with the features returned by our deterministic
feature selection algorithm (same comment applies to Theorem~\ref{thm2}).
The corresponding algorithm is presented as Algorithm~\ref{alg:feature1} along with
the proof of the theorem in Section~\ref{sec:proofs}.
Prior to this result, the best, and in fact the only method with theoretical guarantees for this supervised setting~\cite{BMD09c,BZMD11} is randomized\footnote{We should note that~\cite{BMD09c} describes
the result for the unsupervised setting but it's easy to verify that the same algorithm and bound apply
to the supervised setting as well.} and gives
$$
\cl F(\mathcal{P},\cl S_{out})
\leq \gamma\cdot(3 + O(\sqrt{{k \log(k)}/{r}})) \cdot \cl F(\mathcal{P},\cl S_{in}).
$$
Further,~\cite{BMD09c,BZMD11} requires $r = \Omega( k \log k )$, otherwise the analysis breaks.
We improve this bound by a factor of $O(\log k)$; also, we allow the user to select as few as  $r=O(k)$ features.

A surprising existential result is a direct corollary of the above theorem by setting \math{\cl S_{in} = \cl S_{opt}}
and using an optimal clustering algorithm on the reduced dimension data
($\gamma=1$). 
\begin{corollary}\label{col}
For any set of points $\mathcal{P} = \{ \p_1, \p_2, ..., \p_m \} \in \R^{m\times n}$, integer $k$, and $\epsilon > 0$,
there is a set of $r = O(k / \epsilon^2)$ features $\mathcal{\hat{P}} = \{ \hat{\p}_1, \hat{\p}_2, ..., \hat{\p}_m \} \in \R^{m\times r}$
such that, if
\mand{
\cl{S}_{opt} = \argmin_{\cl{S}} \cl{F} (\mathcal{P}, \cl{S})
\qquad\text{and} \qquad
\hat{\cl{S}}_{opt} = \argmin_{\cl{S}} \cl{F} (\mathcal{\hat{P}}, \cl{S}),
}
then,
$$ \cl{F} (\mathcal{P}, \hat{\cl{S}}_{opt}) \leq \left(5 + \epsilon \right) \cl{F} (\mathcal{P}, \cl{S}_{opt}) .$$
\end{corollary}
In words, for any dataset there exist a small subset
of  $O(k/\epsilon^2)$ features such that the optimal clustering
on these features is at most a $\left(5 + \epsilon \right)$-factor
worse than the optimal clustering
on the original features. Unfortunately,
finding these features without the knowledge of $\cl S_{opt}$
is not obvious.

\subsection{Unsupervised Feature Selection} Our second result is within the context of unsupervised feature selection.
In unsupervised feature selection, the goal is to obtain a $k$-partition that is as close as possible to the optimal $k$-partition of the high dimensional data.
The theorem below shows that it is possible to reduce the dimension of any high-dimensional dataset by using a deterministic method and obtain some theoretical guarantees on the quality of the clusters.
\begin{theorem}\label{thm2}
There is an $O(m n \min\{m,n\} + rk^2n)$ time deterministic algorithm
which takes as input any set of points $\mathcal{P} = \{ \p_1, \p_2, ..., \p_m \} \in \R^{m\times n}$, a number of clusters $k$,
and a number of features to be selected
$k < r < n$. The algorithm
constructs
$\mathcal{\hat{P}} = \{ \hat{\p}_1, \hat{\p}_2, ..., \hat{\p}_m \} \in \R^{m\times r}$ such that \math{\cl S_{out}},
an arbitrary $\gamma$-approximation
on $\mathcal{\hat{P}}$, satisfies
\eqan{
\cl F(\mathcal{P},\cl S_{out})
&\le&
\left(1 + 4 \gamma \frac{( 1 + \sqrt{n/r})^2}{(1 - \sqrt{k/r} )^{2}} \right)
\cl F_{opt}.\\
}
\end{theorem}
Asymptotically, as \math{r\rightarrow\infty} and \math{n/r\rightarrow\infty},
the approximation factor is
\math{\gamma\cdot O\left(n/r\right)}.
The corresponding algorithm is presented as Algorithm~\ref{alg:feature2} along with
the proof of the theorem in Section~\ref{sec:proofs}.
Prior to this result, the best method for this task was given
 in~\cite{BMD09c,BZMD11}, which replaces
$O\left(n/r\right)$ with $\left(3 + O\left(\sqrt{{k \log(k)}/{r}}\right)\right) $ and it is \emph{randomized}, i.e.
this bound is achieved only with a constant probability.
Further,~\cite{BMD09c,BZMD11} requires $r = \Omega( k \log k )$, so one \emph{cannot} select $o\left(k \log k\right)$ features.
Clearly, our bound is worse but we achieve it deterministically, and it applies to any $r > k$, even \math{r=k+1}.

Finally, it is possible to combine
the algorithm of Theorem~\ref{thm2} with the randomized
algorithm of~\cite{BMD09c,BZMD11} and obtain the following result.
\begin{theorem}\label{thm3}
There is an $O\left(m n k + r k^3 \log(k) + r \log r \right)$ time randomized algorithm which takes as input any set of points $\mathcal{P} = \{ \p_1, \p_2, ..., \p_m \} \in \R^{m\times n}$, a
number of clusters $k$, and a number of features to be selected
$k < r < 4 k \log k $. The algorithm
constructs
$\mathcal{\hat{P}} = \{ \hat{\p}_1, \hat{\p}_2, ..., \hat{\p}_m \} \in \R^{m\times r}$ such that \math{\cl S_{out}},
an arbitrary $\gamma$-approximation
on $\mathcal{\hat{P}}$, satisfies with probability $0.4$,
\eqan{
F(\mathcal{P},\cl S_{out}) &\le&
\left( 15 + 320\gamma
\left(\frac{1+\sqrt{{16 k \log(20 k) }/{ r }}}{1-\sqrt{k/r}}\right)^2\right)
 \cl F_{opt}.\\
}
\end{theorem}
Since \math{r<4k\log k}, the approximation factor is
\math{\gamma\cdot O\left(k\log k/r\right)}; there is no result in the
literature which provides an approximation guarantee for selecting
\math{o(k\log k)} features.
Note that the theorem is stated with the assumption \math{r<4 k\log k},
even though the corresponding algorithm (Algorithm~\ref{alg:feature3})
runs with any \math{r>k}.
This is because the algorithm of
Theorem~\ref{thm3} is essentially
the main algorithm of ~\cite{BMD09c,BZMD11} when
\math{r=\Omega(k\log k)} (that is, we do not
provide an improved result for \math{r=\Omega(k\log k)}).
What we achieve is to
break the barrier of having to select
$r = \Omega\left(k \log k\right)$ features. So, Theorem~\ref{thm3},
to the best of our knowledge, is the best available algorithm
in the literature providing theoretical guarantees for
 unsupervised feature selection in
$k$-means clustering using $O\left(k\right)$ features.
The proof of Theorem~\ref{thm3} is given in Section~\ref{sec:proofs}.
Comparing Theorem~\ref{thm3} with
Theorem~\ref{thm2}, we obtain a much better approximation bound
at the cost of introducing randomization.
We note here that the main algorithm of~\cite{BMD09c} can be
 obtained as Algorithm~\ref{alg:feature3} in Section~\ref{sec:proofs}
after ignoring the 4th and 5th steps and replacing the approximate
SVD with the exact SVD in the first step
(the algorithm in~\cite{BZMD11} uses approximate SVD as we do in
Algorithm~\ref{alg:feature3}).

\subsection{Discussion}

Supervised feature selection for \math{k}-means is not prevalent in the
literature, precisely because most of the time one is interested in obtaining
the input partition \math{\cl S_{in}} to begin with. The practical
implication of our result is that
it is possible to identify (efficiently) a small set of important feature
dimensions that are sufficient for essentially reproducing a given clustering.
We also point out that there are randomized algorithms which can quickly
give a good input clustering \math{\matX_{in}} from which to obtain the features;
for example the \math{k}-means\math{{++}} algorithm \cite{AV07} provides
a randomized input clustering that is an \math{O(\log k)}-factor
approximation. Using this clustering will yield randomized feature dimensions such that
clustering in the reduced dimension would also give an
\math{O(\log k)} factor approximation.


For unsupervised feature selection, the theoretical guarantee
from the deterministic algorithm is weak, but we suspect that it may perform
very well in practice on typical input data matrices. The empirical
investigation of this algorithm would also be an interesting future
direction.

All our algorithms require the top $k$ singular vectors  of  the data matrix
(or
an approximation to the top \math{k} singular vectors).
Then, the selection of the features is done by looking at the structure of these singular vectors and using
the deterministic techniques of~\cite{BSS09,BDM11a}. We should note that~\cite{BMD09c,BZMD11} takes a similar approach
as far as computing the (approximate) singular vectors of the dataset; then~\cite{BMD09c,BZMD11} employ
the randomized technique of~\cite{RV07} to extract the features.

The high level description of
 our results for unsupervised feature selection assumed that the
output partition, obtained in the reduced-dimension space,
is the optimal one:
\math{\cl S_{out} = \hat{\cl{S}}_{opt} }.
This is the case if we assume a $\gamma$-approximation
algorithm with $\gamma=1$. Notice, though,
 that our theorems are actually more general and can accomodate
any $\gamma \ge 1$.

\section{Preliminaries}\label{sec:pre}
We now provide the necessary technical
 background that we will use in presenting
our algorithms and proving our main theorems in Section~\ref{sec:proofs}.

\subsection{Singular Value Decomposition}\label{sec:svd}
The Singular
Value Decomposition (SVD)
of a matrix plays an important role in the description of our algorithms.
The SVD of a matrix
$\matA \in \R^{m \times n}$ with $\rank(\matA)=\rho\le\min\{m,n\}$ is
\begin{eqnarray*}
\label{svdA} \matA
         = \underbrace{\left(\begin{array}{cc}
             \matU_{k} & \matU_{\rho-k}
          \end{array}
    \right)}_{\matU_\matA \in \R^{m \times \rho}}
    \underbrace{\left(\begin{array}{cc}
             \matSig_{k} & \bf{0}\\
             \bf{0} & \matSig_{\rho - k}
          \end{array}
    \right)}_{\matSig_{\matA} \in \R^{\rho \times \rho}}
    \underbrace{\left(\begin{array}{c}
             \matV_{k}\transp\\
             \matV_{\rho-k}\transp
          \end{array}
    \right)}_{\matV_\matA\transp \in \R^{\rho \times n}},
\end{eqnarray*}
with singular values \math{\sigma_1\ge\sigma_2\ge\cdots\ge\sigma_\rho>0}
contained in \math{\matSig_k \in \R^{k \times k}} and \math{\matSig_{\rho-k}
\in \R^{(\rho-k) \times (\rho-k)}}.
$\matU_k \in \R^{m \times k}$ and $\matU_{\rho-k} \in \R^{m \times (\rho-k)}$ contain the left singular vectors of $\matA$.
Similarly, $\matV_k \in R^{n \times k}$ and $\matV_{\rho-k} \in \R^{n \times (\rho-k)}$ contain the right singular vectors.
In our description of the SVD, we have explicitly separated the singular
vectors into the top \math{k} singular vectors and the remaining
\math{\rho-k} sincular vectors (where \math{k} is the input number of
clusters); this is because there is a relationship in our algorithms
between the input number of clusters \math{k} and the top-\math{k} singular
vectors.
We use $\matA^+ = \matV_\matA \matSig_\matA^{-1} \matU_\matA\transp \in \R^{n \times m}$
to denote the Moore-Penrose
pseudo-inverse of $\matA$ with $\matSig_\matA^{-1}$ denoting the inverse of
$\matSig_\matA$. We use the Frobenius and the spectral matrix
norms: $ \FNormS{\matA} = \sum_{i,j} \matA_{ij}^2=\sum_{i=1}^\rho\sigma_i^2 $;
and $\TNorm{\matA}^2 = \sigma_1^2$, respectively.
Given $\matA$ and $\matB$ of appropriate dimensions,
$\FNorm{\matA\matB} \leq \FNorm{\matA}
\TNorm{\matB}$;  we call this property spectral submultiplicativity because it
is a stronger version of the standard matrix-norm
submultiplicativity property $\FNorm{\matA\matB} \leq \FNorm{\matA}
\FNorm{\matB}$.
Let $\matA_k=\matU_k \matSig_k \matV_k\transp = \matA \matV_k \matV_k\transp$ and
$\matA_{\rho-k} = \matA - \matA_k = \matU_{\rho-k}\matSig_{\rho-k}\matV_{\rho-k}\transp$.
The SVD gives the best rank-\math{k} approximation to \math{\matA} in both the
spectral and Frobenius norms: if \math{\rank(\tilde\matA)\le k} then (for $\xi=2,\mathrm{F}$)
\math{\XNorm{\matA-\matA_k}\le\XNorm{\matA-\tilde\matA}}; and,
$\FNorm{\matA-\matA_k} = \FNorm{\matSig_{\rho-k}}$.

\subsection{Approximate Singular Value Decomposition}
The exact SVD of $\matA$, though a deterministic algorithm, takes cubic time.
More specifically, for any $k \ge 1$, the running time to compute the top $k$
left and/or right singular vectors of $\matA \in \R^{m \times n}$ is $O(mn\min\{m,n\})$
(See Section 8.6 in~\cite{GV96}).
We will use the exact SVD in our deterministic feature selection algorithms in Theorems~\ref{thm1} and~\ref{thm2}.
To speed up our randomized algorithm in Theorem~\ref{thm3}, we will use a factorization, which can be computed fast
and approximates the SVD in some well defined sense.
We quote a recent result from~\cite{BDM11a} for a relative-error Frobenius norm SVD approximation algorithm (which is proved in \cite{BDM11b}).
The exact description of the corresponding algorithm implied by
Lemma \ref{tropp2}  is out of the scope of this work.
\begin{lemma}
[Lemma 11 in~\cite{BDM11b}]
\label{tropp2}
Given \math{\matA\in\R^{m\times n}} of rank $\rho$, a target rank $2\leq k < \rho$, and $0 < \epsilon < 1$, there exists an
$O\left(mnk/\epsilon\right)$ time randomized algorithm that  computes a matrix $\matZ \in \R^{n \times k}$ and a matrix
$\matE = \matA - \matA \matZ \matZ\transp \in \R^{m \times n}$
such that
$\matZ\transp\matZ = \matI_k$, \math{\matE\matZ=\bm{0}_{m \times k}}, and
$$\Expect{\FNormS{\matE}} \leq \left(1+{\epsilon}\right)\FNormS{\matA - \matA_k}.$$
%
We use $\matZ = \text{FastApproximateSVD}(\matA, k, \epsilon)$ to denote this
randomized procedure.
\end{lemma}
The matrix \math{\matZ} is an approximate version of \math{\matV_k}, hence the
notion fast SVD.

\subsection{Linear Algebraic Definition of $k$-means}\label{sec:linear}
We now give the linear algebraic definition of the $k$-means problem.
Recall that $\mathcal{P} = \{ \p_1, \p_2, ..., \p_m \} \in \R^{m\times n}$ contains the data points,
$k$ is the number of clusters, and \math{\cl S} denotes a $k$-partition of $\mathcal{P}$.
Define the data matrix
\math{\matA \in \R^{m \times n}} as
$$\matA=
\left[\begin{matrix}\text{--- }\p_1\transp\text{ ---}\\\text{--- } \p_2\transp\text{ ---}\\ \vdots\\ \text{--- }\p_m\transp\text{ ---}
\end{matrix}\right].$$
We represent a clustering \math{\cl S} by its cluster indicator matrix
$\matX \in \R^{m \times k}$.
Each column \math{j=1,\ldots,k} of \math{\matX} represents a cluster.
Each row \math{i=1,\ldots,m}
indicates the
cluster membership of the point \math{\p_i}. So,
$$\matX_{ij}=1/\sqrt{s_j},$$ if and
only if the data point \math{\p_i} is in cluster \math{S_j} ($s_j = |\cl S_j|$).
Every row of $\matX$ has one non-zero element,
corresponding to the cluster to which the data
point belongs to. There are \math{s_j} non-zero elements in column
\math{j}, which indicates the points belonging to
\math{S_j}. Obesrve that \math{\matX} is a matrix with orthonormal columns.
We now obtain the matrix formulation of the
\math{k}-means problem:
\vspace{-0.095in}
\eqan{
\cl F(\matA,\matX)
&:=&
\FNormS{\matA - \matX \matX\transp \matA}\\
&=& \sum_{i=1}^{m} \norm{ \p_i\transp - \matX_{i} \matX\transp\matA }_2^2\\
&=& \sum_{i=1}^m\norm{\p_i\transp-\bm\mu(\p_i)\transp}_2^2 \\
&=& \mathcal{F}(\mathcal{P}, \cl S),
}
where we define \math{\matX_i} as the \math{i}th row of \math{\matX} and
we have used the identity $\matX_{i} \matX\transp\matA=\bm\mu(p_i)\transp,$ for $i=1,...,m,$.
This identity
is true because \math{\matX\transp\matA} is a matrix whose row \math{j}
is \math{\sqrt{s_j}\bm\mu_j}, proportional to the centroid of the
\math{j}th cluster; now, \math{\matX_{i}} picks the row corresponding to its
non-zero element, i.e. the cluster corresponding to point \math{i}, and scales
it by \math{1/\sqrt{s_j}}.
Using this formulation, the goal of $k$-means is to find an indicator
matrix $\matX$ which minimizes
\math{\FNormS{\matA - \matX \matX\transp \matA}}.
From now on, \math{\matX\in\R^{m\times k}} will refer generically to such an indicator matrix.
We will denote the best such indicator
 matrix \math{\matX_{opt}}:
$$ \matX_{opt} = \argmin_{\matX \in \R^{m \times k}} \FNormS{ \matA - \matX \matX\transp \matA };$$
so,
$$\cl F_{opt} = \FNormS{\matA - \matX_{opt} \matX_{opt}\transp \matA}.$$
Since \math{\matA_k} is the best
rank \math{k} approximation to \math{\matA}, $\FNormS{ \matA - \matA_k } \le
\mathcal{F}_{opt},$
because $\matX_{opt} \matX_{opt}\transp \matA$ has rank at most $k$.

Using the matrix formulation for $k$-means, we can restate the goal of a feature selection algorithm
in matrix notation. So, the goal of feature selection is to construct points $\matC \in \R^{m \times r}$, where $\matC$
is a subset of $r$ columns from $\matA$, which represent the $m$ points in the $r$-dimensional selected feature space.
Note also that we will allow rescaling of the corresponding columns,
i.e. multiplication of each column with a scalar.
Now consider the optimum $k$-means partition of the points in $\matC$:
$$ \hat{\matX}_{opt} = \argmin_{\matX \in \R^{m \times k}} \FNormS{ \matC - \matX \matX\transp \matC }.$$
(recall that $\matX \in \R^{m \times k}$
is restricted to be an indicator matrix.)
The goal of feature selection is to construct the new set of points $\matC$ such that,
$$ \FNormS{ \matA - \hat{\matX}_{opt} \hat{\matX}_{opt}\transp \matA } \leq \alpha \FNormS{ \matA - \matX_{opt} \matX_{opt}\transp \matA }.$$
More generally, we may use a \math{\gamma}-approximate clustering algorithm
to construct an approximation to \math{\hat\matX_{opt}}.

\subsection{Spectral Sparsification and Rudelson's concentration Lemma}
We now present the main tools we use to select the features in the context of $k$-means clustering.
Let $\matOmega \in \R^{n \times r}$ be a matrix such that $\matA \matOmega \in \R^{m \times r}$
contains $r$ columns of $\matA$. The matrix $\matOmega$ is a ``sampling''
 projection operator onto the $r$-dimensional subset of features.
Let $\matS \in \R^{r \times r}$ be a diagonal matrix, so
$\matA \matOmega \matS \in \R^{m \times r}$ rescales the columns of $\matA$ that are in $\matA \matOmega$.
Intuitively, $\matA \matOmega \matS$ projects down to the chosen $r$ dimensions and then rescales the data points along these dimensions.

In this section we present two deterministic and a randomized algorithm for constructing such matrices $\matOmega$ and $\matS$.
All three algorithms come from prior work but we include a full description of them for completeness.
The corresponding lemmas in this subsection describe certain spectral properties of these matrices.
The proofs of these Lemmas appeared in prior work, so we only give appropriate references and do
not prove the results here.

\begin{algorithm}[t] \label{alg:sample1}
\begin{framed}
   \caption{{\sf DeterministicSamplingI} (Lemma~\ref{theorem:3setGeneralF})}
   {\bfseries Input:} $\matV\transp=[\v_1, \v_2, \ldots,\v_n] \in \R^{k \times n}$,
\math{\matB=[\b_1, \b_2, \ldots,\b_n] \in \R^{\ell_1 \times n}},
and \math{r > k}. \\
    {\bfseries Output:} Sampling matrix $\matOmega \in \R^{n \times r}$ and rescaling matrix $\matS \in \R^{r \times r}$.
 \begin{algorithmic}[1]
    \STATE Initialize  $\matA_0 = \bm{0}_{k \times k}$, $\matOmega = \bm{0}_{n \times r}$, and $\matS=\bm{0}_{r \times r}$.
    \STATE Set constants $
\delta_\matB=\FNormS{\matB} (1-\sqrt{k/r})^{-1};\
\delta_L=1$.
   \FOR{$\tau=0$ {\bfseries to} $r-1$}
   \STATE Let \math{\scl_\tau=\tau-\sqrt{rk}}.
   \STATE Pick index $i_{\tau}\in\{1,2,...,n\}$ and number $t_{\tau}>0$ (see text for the definition of $U, L$):
    $$\hspace*{-0.2in} U(\b_{i_{\tau}},\delta_\matB) \le\frac{1}{t_{\tau}}\le
       L(\v_{i_{\tau}}, \delta_L, \matA_{\tau},\scl_\tau).$$
   \STATE Update $\matA_{\tau+1} = \matA_{\tau} + t_{\tau} \v_{i_{\tau}} \v_{i_{\tau}}\transp$; set $\matOmega_{{i_{\tau}},\tau+1} = 1$ and $\matS_{\tau+1,\tau+1} = 1/\sqrt{t_{\tau}}$.
   \ENDFOR
   \STATE Multiply all the weights in $\matS$ by
\mand{\sqrt{ r^{-1}(1-\sqrt{k/r}) }.}
   \STATE {\bfseries Return:} $\matOmega$ and $\matS$.
\end{algorithmic}
\end{framed}
\end{algorithm}
\begin{algorithm}[t] \label{alg:sample2}
\begin{framed}
   \caption{{\sf DeterministicSamplingII} (Lemma~\ref{theorem:3setGeneralS})}
   {\bfseries Input:} $\matV\transp=[\v_1, \v_2, \ldots,\v_n] \in \R^{k \times n}$,  \math{\matQ=[\q_1, \q_2,\ldots,\q_d] \in \R^{\ell_2 \times n}},
and \math{r > k}. \\
    {\bfseries Output:} Sampling matrix $\matOmega \in \R^{n \times r}$ and rescaling matrix $\matS \in \R^{r \times r}$.
 \begin{algorithmic}[1]
    \STATE Initialize  $\matA_0 = \bm{0}_{k \times k}$, $\matB_0 = \bm{0}_{\ell_2 \times \ell_2}$, $\matOmega = \bm{0}_{n \times r}$, and $\matS=\bm{0}_{r \times r}$.
    \STATE Set constants $
\delta_\matQ=\left( 1 + \ell_2/r \right) \left(1-\sqrt{k/r}\right)^{-1};\ \delta_L=1$.
   \FOR{$\tau=0$ {\bfseries to} $r-1$}
   \STATE Let \math{\scl_\tau=\tau-\sqrt{rk}}; 
   \math{\scu_{\tau} = \delta_{\matQ} \left( \tau + \sqrt{\ell_2 r} \right) }
   \STATE Pick index $i_{\tau}\in\{1,2,...,n\}$ and number $t_{\tau}>0$ (see text for the definition of $U, L$):
    $$\hspace*{-0.2in} \hat{U}(\q_{i_{\tau}},\delta_\matQ,\matB_{\tau},\scu_\tau) \le\frac{1}{t_{\tau}}\le
       L(\v_{i_{\tau}}, \delta_L, \matA_{\tau},\scl_\tau).$$
   \STATE Update $\matA_{\tau+1} = \matA_{\tau} + t_{\tau} \v_{i_{\tau}} \v_{i_{\tau}}\transp$;
                 $\matB_{\tau+1} = \matB_{\tau} + t_{\tau} \q_{i_{\tau}} \q_{i_{\tau}}\transp$, and \\
   set $\matOmega_{{i_{\tau}},\tau+1} = 1$, $\matS_{\tau+1,\tau+1} = 1/\sqrt{t_{\tau}}$.
   \ENDFOR
   \STATE Multiply all the weights in $\matS$ by $ \sqrt{ r^{-1} \left(1-\sqrt{k/r}\right) }.$
   \STATE {\bfseries Return:} $\matOmega$ and $\matS$.
\end{algorithmic}
\end{framed}
\end{algorithm}

\begin{lemma} [Lemma 13 in~\cite{BDM11b}]
\label{theorem:3setGeneralF}
Let \math{\matV\transp \in \R^{k \times n}} and
\math{\matB \in \R^{\ell_1 \times n}} with \math{\matV\transp\matV=\matI_k}. Let $r > k$.
Algorithm~\ref{alg:sample1} runs in $O(rk^2n + \ell_1n )$ time and deterministically constructs a sampling matrix \math{\matOmega\in\R^{n \times r}} and
a rescaling matrix \math{\matS\in \R^{r \times r}} such that,
\begin{align*}
&\sigma_k(\matV\transp \matOmega \matS) \ge 1 - \sqrt{{k}/{r}};
&\FNorm{\matB \matOmega \matS}         \le   \FNorm{\matB}.
\end{align*}
\end{lemma}

Algorithm \ref{alg:sample1} is a greedy technique that selects columns one at a time.
To describe the algorithm in more detail, it is convenient to view the input matrices as two sets of \math{n} vectors,
\math{\matV\transp=[\v_1, \v_2, \ldots,\v_n]} and
\math{\matB=[\b_1, \b_2, \ldots,\b_n]}.
Given  $k$ and $r>k$, introduce the
iterator $\tau = 0, 1,2,...,r-1,$ and define the
parameter $\scl_\tau=\tau-\sqrt{rk}$.
For a square symmetric
matrix \math{\matA\in\R^{k\times k}} with eigenvalues \math{\lambda_1,\ldots,
\lambda_k}, $\v \in \R^k$ and $\scl\in\R$, define
$$
\phi(\scl, \matA) =
\sum_{i=1}^k\frac{1}{\lambda_i-\scl},
$$
and let
$L(\v, \delta_L, \matA, \scl)$  be defined as
$$ L(\v, \delta_L, \matA, \scl)  =
\frac{\v\transp(\matA-\scl'\matI_k)^{-2}\v}
{\phi(\scl', \matA)-\phi(\scl,\matA)} -\v\transp(\matA-\scl'\matI_k)^{-1}\v,$$
where $$\scl'= \scl+\delta_L = \scl+1.$$
For a vector $\z$ and scalar \math{\delta > 0}, define the function
$$
U(\z,\delta) = \frac{1}{\delta} \z\transp \z .
$$
At each iteration \math{\tau}, the algorithm selects
$i_{\tau}$, $t_{\tau} > 0$
for which
$$U(\b_{i_{\tau}},\delta_\matB)\le t_{\tau}^{-1} \le
       L(\v_{i_{\tau}},\delta_L,\matA_{\tau},\scl_\tau).$$
The running time of the algorithm
is dominated by the search for an index $i_{\tau}$
satisfying $$U(\b_{i_{\tau}},\delta_\matB)\le t_{\tau}^{-1}\le
       L(\v_{i_{\tau}},\delta^{-1},\matA_{\tau},\scl_\tau)$$ (one can achieve that by exhaustive search). One needs
\math{\phi(\scl,\matA)}, and hence the eigenvalues of
\math{\matA}. This takes
\math{O(k^3)} time, once per iteration, for a total of
\math{O(rk^3)}. Then, for \math{i=1,\ldots,n}, we need to compute
\math{L} for every \math{\v_i}. This takes \math{O(nk^2)} per iteration,
for a total of \math{O(rnk^2)}. To compute \math{U}, we need
\math{\b_i\transp\b_i}  for \math{i=1,\ldots,n},
which need to be computed only once for the whole algorithm and
takes \math{O(\ell_1 n )}.
So, the total running time is \math{O(nrk^2+\ell_1 n)}.
\begin{lemma} [Lemma 12 in~\cite{BDM11b}]
\label{theorem:3setGeneralS}
Let \math{\matV\transp \in \R^{k \times n}}, \math{\matQ \in
\R^{\ell_2 \times n} }, \math{\matV\transp\matV=\matI_k}, and
\math{\matQ\transp\matQ=\matI_{\ell_2}}. Let $r > k$.
Algorithm~\ref{alg:sample2} runs in  $O( r k^2 n + r \ell_2^2n )$ time and deterministically constructs a sampling matrix
\math{\matOmega\in\R^{n \times r}} and
a rescaling matrix \math{\matS\in \R^{r \times r}} such that,
\begin{align*}
&\sigma_k(\matV\transp \matOmega \matS) \ge 1 - \sqrt{{k}/{r}};
&\TNorm{\matQ \matOmega \matS}          \le  1+\sqrt{\ell_2/r}.
\end{align*}
Moreover, if $\matQ = \matI_{n}$, the running time is $O( r k^2 n )$.
\end{lemma}
Algorithm \ref{alg:sample2} is similar to Algorithm \ref{alg:sample1}; we only need to define the function
$\hat{U}$.
For a square symmetric matrix \math{\matB\in\R^{\ell_2 \times \ell_2}} with eigenvalues \math{\lambda_1,\ldots,
\lambda_{\ell_2}}, $\q \in \R^{\ell_2}$, $\scu\in\R$, define:
$$
\hat{\phi}(\scu,\matB) =
\sum_{i=1}^{\ell_2}\frac{1}{\scu - \lambda_i},
$$
and let
$\hat{U}(\q, \delta_{\matQ}, \matB, \scu)$  be defined as
$$ \hat{U}(\q, \delta_{\matQ}, \matB, \scu)  =
\frac{\q\transp(\matB-\scu'\matI_{\ell_2})^{-2}\q}
{\hat\phi(\scu,\matB)-\hat{\phi}(\scu', \matB) }
-
\q\transp(\matB-\scu'\matI_{\ell_2})^{-1}\q,$$
where
$$ \scu' = \scu + \delta_{\matQ} =
\scu + \left( 1 + \ell_2/r \right) \left(1-\sqrt{k/r}\right)^{-1}
.$$
The running time of the algorithm is \math{O(nrk^2+ nr\ell_2^2)}.

Lemmas~\ref{theorem:3setGeneralF} and~\ref{theorem:3setGeneralS} are generalizations
of the original work of Batson et al.~\cite{BSS09}, which presented a deterministic
algorithm that operates only on $\matV$. Lemmas~\ref{theorem:3setGeneralF} and~\ref{theorem:3setGeneralS} are
proved in~\cite{BDM11b} (see Lemmas 12 and 13 in~\cite{BDM11b}).
We will use Lemma~\ref{theorem:3setGeneralF} in a novel way:
we will apply it to a matrix $\matB$ of the form
$$\matB = \left(\begin{array}{c}
             \matB_1\\
             \matB_2
          \end{array}
    \right)
,$$
so we will be able to control the sum of the Frobenius norms of two different matrices $\matB_1$, $\matB_2$, which is
all we need in our application. In the above formula, $\matB$ is a $2 \times 1$ block matrix with $\matB_1$ and $\matB_2$
being the underlying blocks. The above two lemmas will be used to prove our deterministic results for feature selection,
i.e. Theorems~\ref{thm1} and~\ref{thm2}.
\begin{algorithm}[t]\label{alg:sample3}
\begin{framed}
   \caption{{\sf RandomizedSampling (Lemma~\ref{theorem:3setGeneralFr})}}
   {\bfseries Input:} $\matV\transp=[\v_1, \v_2, \ldots,\v_n] \in \R^{k \times n}$,
and the number of sampled columns \math{r > 4 k \log k}. \\
    {\bfseries Output:} Sampling matrix $\matOmega \in \R^{n \times r}$ and rescaling matrix $\matS \in \R^{r \times r}$.
\begin{algorithmic}[1]
   \STATE For $i = 1, ..., n$ compute  $ p_i = \frac{1}{k}\norm{\v_i}^2_2.$
   \STATE Initialize $\matOmega = \bm{0}_{n \times r}$ and $\matS=\bm{0}_{r \times r}$.
   \FOR{$\tau=1$ {\bfseries to} $r$}
   \STATE Select index $i\in\{1,2,...,n\}$ independently with
the probability of selecting index \math{i} equal to  $p_i$.
   \STATE Set $\matOmega_{i,\tau} = 1$ and $\matS_{\tau,\tau} = 1/\sqrt{p_i r}$.
   \ENDFOR
   \STATE {\bfseries Return:} $\matOmega$ and $\matS$.
\end{algorithmic}
\end{framed}
\end{algorithm}

We will also need the following result, which corresponds to the celebrated work of Rudelson and Vershynin~\cite{Rud99,RV07}
and describes a randomized algorithm for constructing matrices $\matOmega$ and $\matS$. The lower bound with the optimal
constants $4$ and $20$ was obtained more recently as Lemma 15 in~\cite{Mal10}.
The Frobenius norm bounds are straightforward; a short proof can be found as Eqn.~36 in~\cite{DMM08}.
This lemma will be used to prove our hybrid randomized result for feature selection, i.e. Theorem~\ref{thm3}.
\begin{lemma}
\label{theorem:3setGeneralFr}
Let \math{\matV\transp \in \R^{k \times n}}, \math{\matB \in \R^{\ell_1 \times n}} and
\math{\matQ \in \R^{\ell_2 \times n}}, with \math{\matV\transp\matV=\matI_k}. Let $r > 4 k \log k$.
Algorithm \ref{alg:sample3}, in $O(nk + r\log r)$ time,  constructs a sampling
matrix  \math{\matOmega\in\R^{n \times r}} and a rescaling matrix
\math{\matS\in \R^{r \times r}} such that
\begin{align*}
& \Expect{\norm{\matQ \matOmega \matS}_\mathrm{F}^2}          =  \norm{\matQ}_\mathrm{F}^2;
&\Expect{\norm{\matB \matOmega \matS}_\mathrm{F}^2 }       =   \norm{\matB}_\mathrm{F}^2; \end{align*}
and with probability at least  $0.9$,
$$
\sigma_k^2(\matV\transp \matOmega \matS) \ge 1 - \sqrt{{4 k \log(20 k )}/{r}}.
$$
\end{lemma}

\section{A structural bound for clustering error  with feature selection}\label{sec:algs}

In this section, we develop two lemmas which are
general structural bounds for the clustering error when
using feature selection. These lemmas are the
basis for all of our algorithms, and hence are crucial to proving
our results, specifically
Theorems~\ref{thm1}, \ref{thm2}, and~\ref{thm3}
(recall that there is no algorithm for Corollary~\ref{col} since that
 result is
non-constructive).

In the following two lemmas, the sampling and rescaling matrices
$\matOmega \in \R^{n \times r}$ and $\matS \in \R^{r \times r}$
 are \emph{arbitrary}, modulo the rank restriction in the lemmas.
Thus, one can apply the lemmas to
 a general dimension reduction matrix $\matPsi = \matOmega \matS$.

\begin{lemma}
\label{lem:generic2}
Let
$\matA \in \R^{m \times n}$ be the input data matrix,
$k > 0$ an integer,
and $\matX_{in} \in \R^{m \times k}$ an input \math{k}-clustering
indicator matrix.
Let $\matOmega \in \R^{n \times r}$ and $\matS \in \R^{r \times r}$ be any matrices, and set
 $\matC = \matA \matOmega \matS \in \R^{m \times r}$. Let $\matZ \in \R^{n \times k}$ be any orthonormal matrix, and define the residual \math{\matE} by
 $\matA = \matA \matZ \matZ\transp + \matE$, where the matrix
 $\matE \in \R^{m \times n}$.
Let $\matX_{out}$ be the output of some $\gamma$-approximation algorithm on
($\matC$, $k$). If
$\rank(\matZ\transp
\matOmega\matS)=k$, then
\mand{
\FNormS{\matA-\matX_{out}\matX_{out}\transp\matA}\\
\le
\FNormS{ \matE}+2\gamma \frac{ \norm{(\matI_{n} - \matX_{in}\matX_{in}\transp)\matA
 \matOmega \matS}_\mathrm{F}^2+\FNormS{\matE\matOmega \matS }}
{\sigma_k^2(\matZ\transp \matOmega \matS)}.
}
\end{lemma}
This lemma bounds the clustering error of the output partition
\math{\matX_{out}} that is constructed in the reduced dimension space.
We get to pick \math{\matZ\matZ\transp}, a projection matrix; if we
project \math{\matA} on  the right using \math{\matZ\matZ\transp}, there
is some residual \math{\matE}. If this residual \math{\matE}
is small (directly
controlled by \math{\matZ}) \emph{and} if the sampling and rescaling matrices
\math{\matOmega,\ \matS} are chosen so that:
(\rn{1}) the input partition has a reasonably small clustering error in the
reduced dimension space;
(\rn{2}) the size of the residual \math{\matE} does not increase significantly
under sampling and rescaling; and
(\rn{3}) the sampling and rescaling does not significantly alter the singular
structure of the projector \math{\matZ},
then the output clustering error is small.
These are the basic three guidelines we need to follow in selecting
\math{\matOmega} and \math{\matS} in our algorithms.

\begin{proof}
We start by manipulating the term
$\norm{\matA -\matX_{out} \matX_{out}\transp \matA}_\mathrm{F}^2$.
First, observe that
$\matA = \matA\matZ\matZ\transp + \matE.$
Next, from the Pythagorean Theorem for matrices\footnote{
Let \math{\matY_1,\matY_2\in\R^{m\times n}} satisfy \math{\matY_1\matY_2\transp=\bm{0}_{m \times m}}. Then,
$
\FNorm{\matY_1+\matY_2}^2 = \FNorm{\matY_1}^2+\FNorm{\matY_2}^2.
$
}
and using that $\matI_{m} - \matX_{out} \matX_{out}\transp$
is a projection matrix we obtain,
\begin{eqnarray*}
\FNormS{\matA -\matX_{out} \matX_{out}\transp \matA}
&=&\FNormS{(\matI_m -\matX_{out} \matX_{out}\transp) \matA\matZ\matZ\transp+
(\matI_m -\matX_{out} \matX_{out}\transp)\matE}\\
&=&\FNormS{(\matI_m -\matX_{out} \matX_{out}\transp) \matA\matZ\matZ\transp} +
\FNormS{(\matI_m -\matX_{out} \matX_{out}\transp)\matE}\\
&\le&
 \FNormS{(\matI_m -\matX_{out} \matX_{out}\transp) \matA\matZ\matZ\transp}
+\FNormS{\matE}.
\end{eqnarray*}
We now bound the first term in the last expression.
Given \math{\matOmega} and \math{\matS},
for some residual matrix $\matY \in \R^{m \times n}$, let
$$\matA\matZ\matZ\transp = \matA \matOmega \matS (\matZ\transp \matOmega \matS)^+
\matZ\transp + \matY.$$
(See
Section~\ref{sec:svd} for the definition of the pseudo-inverse operator.)
Then,
\begin{eqnarray*}
\FNormS{(\matI_m -\matX_{out} \matX_{out}\transp) \matA\matZ\matZ\transp}
&{\buildrel(a)\over\le}&
2 \FNormS{(\matI_{m} -\matX_{out}\matX_{out}\transp)
\matA \matOmega \matS (\matZ\transp \matOmega \matS)^+\matZ\transp} + 2\FNormS{\matY}\\
&{\buildrel(b)\over\le}&
2 \FNormS{(\matI_{m} - \matX_{out}\matX_{out}\transp)\matA \matOmega \matS}\cdot
\TNormS{(\matZ\transp \matOmega \matS)^+} + 2 \FNormS{\matY} \\
&{\buildrel(c)\over\le}&
2 \gamma \FNormS{(\matI_{m} - \matX_{in}\matX_{in}\transp)\matA
 \matOmega \matS}\cdot
\TNormS{(\matZ\transp \matOmega \matS)^+}  + 2 \FNormS{\matY}.
\end{eqnarray*}
In (a), we used \math{\FNormS{\matY_1+\matY_2} \le 2\FNormS{\matY_1} + 2 \FNormS{\matY_2}}
(for any two matrices $\matY_1, \matY_2$),
which follows from the triangle inequality of matrix norms; further we have removed
the projection matrix $\matI_m -\matX_{out}\matX_{out}\transp$ from the second term, which
can be done without increasing the Frobenius norm.
In (b), we used spectral submultiplicativity and the fact that $\matZ\transp$
is orthonormal, and so it can be dropped without increasing the spectral norm.
Finally, in (c), we replaced $\matX_{out}$ by $\matX_{in}$ and the factor $\gamma$ appeared
in the first term. To understand why this can be done,
notice that, by assumption,
$\matX_{out}$ was constructed by running the $\gamma$-approximation on
$\matC = \matA \matOmega \matS$.
So, for any indicator matrix \math{\matX}:
\mand{
\FNormS{(\matI_{m} - \matX_{out}\matX_{out}\transp)\matA \matOmega \matS}
\le
\gamma\FNormS{(\matI_{m} - \matX\matX\transp)\matA \matOmega \matS},
}
and we can set $\matX = \matX_{in}$.
Finally, we bound the term $\FNormS{\matY}$. Recall that
\eqan{
\matY&=&\matA\matZ\matZ\transp-\matA \matOmega \matS (\matZ\transp \matOmega \matS)^+
\matZ\transp\\
&=&
\matA\matZ\matZ\transp-\matA\matZ\matZ\transp \matOmega \matS (\matZ\transp \matOmega \matS)^+
\matZ\transp-(\matA-\matA\matZ\matZ\transp)\matOmega \matS (\matZ\transp \matOmega \matS)^+
\matZ\transp.
}
Note that
$$\matA\matZ\matZ\transp-\matA\matZ\matZ\transp \matOmega \matS (\matZ\transp \matOmega \matS)^+
\matZ\transp = {\bf 0}_{m \times n},$$
since
$\rank(\matZ\transp \matOmega \matS)=k,$ and so
$$\matZ\transp
\matOmega \matS (\matZ\transp \matOmega \matS)^+=\matI_k.$$
So,
\begin{eqnarray*}
\FNormS{\matY}
&=& \FNormS{(\matA-\matA\matZ\matZ\transp)\matOmega \matS (\matZ\transp \matOmega \matS)^+
\matZ\transp}\\
&\leq& \FNormS{(\matA-\matA\matZ\matZ\transp)\matOmega \matS }
\TNormS{ (\matZ\transp \matOmega \matS)^+\matZ}\\
&\leq& \FNormS{(\matA-\matA\matZ\matZ\transp)\matOmega \matS }
\TNormS{ (\matZ\transp \matOmega \matS)^+}\\
&=& \frac{\FNormS{(\matA-\matA\matZ\matZ\transp)\matOmega \matS }}{
\sigma_k^2(\matZ\transp \matOmega \matS)}.\\
\end{eqnarray*}
In the first 3 steps, we have used spectral submultiplicativity, and in
the last step we have used the definition of the
spectral norm of the pseudo-inverse.
Combining all these bounds together (and using $\gamma \geq 1$):
\mand{
\FNormS{\matA-\matX_{out}\matX_{out}\transp\matA}\\
\le
\FNormS{ \matE}+2\gamma \frac{ \norm{(\matI_{n} - \matX_{in}\matX_{in}\transp)\matA
 \matOmega \matS}_\mathrm{F}^2+\FNormS{(\matA-\matA\matZ\matZ\transp)\matOmega \matS }}
{\sigma_k^2(\matZ\transp \matOmega \matS)}.
}

\end{proof}

Lemma~\ref{lem:generic} is a simple corollary of Lemma~\ref{lem:generic2} by using $\matZ = \matV_k \in \R^{n \times k}$,
i.e. the matrix containing the top $k$ right singular vectors of $\matA$. Notice also that in this case $\matE=  \matA - \matA\matV_k\matV_k\transp= \matA-\matA_k$.
Observe that
$$\FNormS{\matE}=\FNormS{\matA - \matA_k} \le \FNormS{ \matA - \matX_{opt} \matX_{opt}\transp \matA}
\le \FNormS{ \matA - \matX_{in} \matX_{in}\transp \matA},$$
The first inequality is from
the optimality of $\matA_k$ (see also Section~\ref{sec:linear});
and, the second inequality is from the  optimality
of the indicator matrix $\matX_{opt}$  for clustering the rows of \math{\matA}
(the high dimensional points).
\begin{lemma}
\label{lem:generic}
Let $\matA \in \R^{m \times n}$, $k > 0$,
 and \math{\matV_k} its top-\math{k} right
singular matrix. Let $\matX_{in} \in \R^{m \times k}$ be an input clustering
indicator matrix and
\math{\matE=\matA-\matA\matV_k\matV_k\transp=\matA-\matA_k}.
Let $\matOmega \in \R^{n \times r}$ and $\matS \in \R^{r \times r}$ be any matrices and set $\matC = \matA \matOmega \matS \in \R^{m \times r}$.
Let $\matX_{out}$ be the output of some $\gamma$-approximation algorithm
on $(\matC, k)$. If
\math{\rank(\matV_k\transp
\matOmega\matS)=k}, then
\eqan{
\FNormS{ \matA -  \matX_{out} \matX_{out}\transp \matA}
&\leq& \FNormS{ \matA -  \matX_{in} \matX_{in}\transp \matA}+
2 \gamma  \frac{ \FNormS{(\matA -\matX_{in} \matX_{in}\transp \matA) \matOmega\matS}  +
\FNormS{\matE  \matOmega\matS}} { \sigma_k^2(\matV_k\transp \matOmega \matS) }
.
}
\end{lemma}

\section{Proofs of Main Theorems}\label{sec:proofs}

\begin{algorithm}[t] \label{alg:feature1}
\begin{framed}
\caption{{\sf Supervised Feature Selection} (Theorem~\ref{thm1})  }
{\bf Input:} \math{\matA \in \R^{m \times n}}, $\matX_{in} \in \R^{m \times k}$, number of clusters $k$,
and number of features \math{r > k}. \\
{\bf Output:} \math{\matC \in \R^{m \times r}} containing $r$ rescaled columns of $\matA$.
\begin{algorithmic}[1]
\STATE Compute the matrix \math{\matV_k \in \R^{n \times k}} from the SVD of $\matA$.
\STATE Let $\matB = \left(\begin{array}{c}
             \matA - \matA \matV_k \matV_k\transp\\
             \matA - \matX_{in}\matX_{in}\transp\matA
          \end{array}
    \right) \in \R^{2m \times n}$; $\matA - \matA \matV_k \matV_k\transp$, $\matA - \matX_{in}\matX_{in}\transp\matA \in \R^{m \times n}$.
\STATE Let $[\matOmega,\matS]={\sf DeterministicSamplingI}(\matV_k\transp, \matB, r). $ (see Algorithm~\ref{alg:sample1})
\STATE {\bf return} \math{\matC=\matA\matOmega\matS \in \R^{m \times r}.}
\end{algorithmic}
\end{framed}
\end{algorithm}

\subsection{Proof of Theorem \ref{thm1}}
We state the corresponding algorithm as Algorithm~\ref{alg:feature1}.
Theorem \ref{thm1} will follow by using Lemmas~\ref{lem:generic} and \ref{theorem:3setGeneralF}. We would like to apply
Lemma~\ref{lem:generic} for the matrices $\matOmega$ and $\matS$ constructed with $DeterministicSamplingI$, i.e. Algorithm~\ref{alg:sample1}.
To do that, we need $$\rank(\matV_k\transp\matOmega\matS)=k.$$ This rank requirement follows from Lemma~\ref{theorem:3setGeneralF},
because $$\sigma_k(\matV_k\transp \matOmega \matS) > 1-\sqrt{k/r} > 0.$$
Hence, Lemma~\ref{lem:generic} gives
\mld{
\FNormS{ \matA -  \matX_{out} \matX_{out}\transp \matA}
\leq \FNormS{ \matA -  \matX_{in} \matX_{in}\transp \matA}+2 \gamma  \frac{ \FNormS{(\matA -\matX_{in} \matX_{in}\transp \matA) \matOmega\matS}  +
\FNormS{\matE  \matOmega\matS}} { \sigma_k^2(\matV_k\transp \matOmega \matS) }
.\label{eq:thm1-1}
}
We can now use the bound for $\sigma_k^2(\matV_k\transp \matOmega \matS)$ from Lemma~\ref{theorem:3setGeneralF}:
\mld{
\sigma^2_k(\matV_k\transp \matOmega \matS) > \left(1-\sqrt{k/r}\right)^2.
\label{eq:thm1-2}
}
Also, we can use the Frobenius norm
bound from Lemma~\ref{theorem:3setGeneralF}.
Our choice of the matrix $\matB\in\R^{2m \times n}$ is
$$\matB = \left(\begin{array}{c}
             \matA - \matA \matV_k \matV_k\transp\\
             \matA - \matX_{in}\matX_{in}\transp\matA
          \end{array}
    \right)=\left(\begin{array}{c}
             \matE\\
             \matA - \matX_{in}\matX_{in}\transp\matA
          \end{array}
    \right).
$$
This bound from Lemma~\ref{theorem:3setGeneralF} gives
\math{\FNormS{\matB \matOmega\matS} \le \FNormS{\matB}}, where,
from our choice of $\matB$,
\eqan{\FNormS{\matB\matOmega\matS}&=&
\FNormS{\matE\matOmega\matS} + \FNormS{(\matA-\matX_{in}\matX_{in}\transp\matA)\matOmega\matS};\\
\FNormS{\matB}& =&
 \FNormS{\matE} + \FNormS{\matA-\matX_{in}\matX_{in}\transp\matA}.}
so, the result of applying  Lemma~\ref{theorem:3setGeneralF} is
\mld{
\FNormS{\matE\matOmega\matS} +
\FNormS{(\matA-\matX_{in}\matX_{in}\transp\matA)\matOmega\matS} \le
\FNormS{\matE} +
\FNormS{\matA-\matX_{in}\matX_{in}\transp\matA}.
\label{eq:thm1-3}
}
Using \r{eq:thm1-2} and \r{eq:thm1-3} in \r{eq:thm1-1},
we have
\mand{
\FNormS{ \matA -  \matX_{out} \matX_{out}\transp \matA}
\leq \FNormS{ \matA -  \matX_{in} \matX_{in}\transp \matA}+\frac{2 \gamma}{(1-\sqrt{k/r})^2} \left( \FNormS{\matA -\matX_{in} \matX_{in}\transp \matA}  +
\FNormS{\matE}\right).
}
Since
\math{\FNormS{\matE}=\FNormS{ \matA - \matA_k }
\le \FNormS{ \matA -  \matX_{in} \matX_{in}\transp \matA},} the result follows.

Lastly, we compute the running time of the algorithm. Algorithm \ref{alg:feature1} computes the matrix $\matV_k$
in $O(m n \min\{m,n\})$ time; then, $ \matA - \matA \matV_k \matV_k\transp$ and $\matA - \matX_{in}\matX_{in}\transp\matA$
can be computed in $O(mnk)$. $DeterministicSamplingI$ takes time $O(rk^2n + m n )$, from
Lemma~\ref{theorem:3setGeneralF}. Overall, the running time of Algorithm \ref{alg:feature1} in Theorem~\ref{thm1}
is $O(m n \min\{m,n\} + rk^2n)$.

\begin{algorithm}[t] \label{alg:feature2}
\begin{framed}
\caption{{\sf Unsupervised Feature Selection}  (Theorem~\ref{thm2}) }
{\bf Input:} \math{\matA \in \R^{m \times n}}, number of clusters $k$,
and number of features \math{r > k}. \\
{\bf Output:} \math{\matC \in \R^{m \times r}} containing $r$ rescaled columns of $\matA$.
\begin{algorithmic}[1]
\STATE Compute the matrix \math{\matV_k \in \R^{n \times k}} from the SVD of $\matA$.
\STATE Let $[\matOmega,\matS]={\sf DeterministicSamplingII}(\matV_k\transp,
\matI_n, r). $ (see Algorithm~\ref{alg:sample2})
\STATE {\bf return} \math{\matC=\matA\matOmega\matS \in \R^{m \times r}.}
\end{algorithmic}
\end{framed}
\end{algorithm}

\subsection{Proof of Theorem \ref{thm2}}
We state the corresponding algorithm as Algorithm~\ref{alg:feature2}.
Theorem \ref{thm2} follows by
combining Lemmas \ref{lem:generic} and \ref{theorem:3setGeneralS}. The rank requirement
in Lemma~\ref{lem:generic} is satisfied by the bound for the smallest singular value of $\matV_k\transp\matOmega\matS$
in Lemma~\ref{theorem:3setGeneralS}.
We are going to apply Lemma~\ref{lem:generic},
with
$$\matX_{in} = \matX_{opt}.$$
Even though we apply Lemma ~\ref{lem:generic} with
$\matX_{in} = \matX_{opt},$ this is merely an artifact of the proof that is
needed to obtain the final result; we do not
actually need to compute $\matX_{opt},$ as is evident in the description of
Algorithm~\ref{alg:feature2}, which runs without knowledge of
$\matX_{opt}.$
We have, from Lemma ~\ref{lem:generic},
\mld{
\FNormS{ \matA -  \matX_{out} \matX_{out}\transp \matA} \leq \FNormS{ \matA -  \matX_{opt} \matX_{opt}\transp \matA}
 + 2 \gamma  \frac{ \FNormS{(\matA -\matX_{opt} \matX_{opt}\transp \matA) \matOmega\matS}  +
\FNormS{\matE  \matOmega\matS}} { \sigma_k^2(\matV_k\transp \matOmega \matS) }
.\label{eq:thm2-1}
}
To bound the second term on the right hand side, we use
spectral submultiplicativity to obtain
$$\FNormS{(\matA -\matX_{opt} \matX_{opt}\transp \matA) \matOmega\matS} \le
\FNormS{(\matA -\matX_{opt} \matX_{opt}\transp \matA)}\cdot \TNormS{ \matI_n \matOmega\matS},$$ and
\eqan{
\FNormS{\matE \matOmega\matS}
&\le &
\FNormS{\matE}\cdot \TNormS{ \matI_n \matOmega\matS}\\ &\le&
\FNormS{(\matA -\matX_{opt} \matX_{opt}\transp \matA)}\cdot \TNormS{ \matI_n \matOmega\matS}.
}
where the last inequality follows because
$\FNormS{ \matE} = \FNormS{\matA-\matA_k} \le
\FNormS{ \matA - \matX_{opt} \matX_{opt}\transp \matA }$.
Plugging back into \r{eq:thm2-1},
\mand{
\FNormS{ \matA -  \matX_{out} \matX_{out}\transp \matA} \leq \FNormS{ \matA -  \matX_{opt} \matX_{opt}\transp \matA}
 + 4 \gamma \FNormS{\matA -\matX_{opt} \matX_{opt}\transp \matA}
\frac{ \TNormS{\matI_n\matOmega\matS}} { \sigma_k^2(\matV_k\transp \matOmega \matS) }
.
}
The result now follows by using the
bounds from Lemma~\ref{theorem:3setGeneralS},
\eqan{
\sigma_k(\matV_k\transp \matOmega \matS)  &\ge &
1 - \sqrt{{k}/{r}}; \\
\TNorm{\matI_n \matOmega \matS}         & \le&  1+\sqrt{n/r}.
}

Finally, we comment on the running time of the algorithm. Algorithm \ref{alg:feature2} computes the matrix $\matV_k$
in $O(m n \min\{m,n\})$ time. $DeterministicSamplingII$ takes time $O(rk^2n )$, from
Lemma~\ref{theorem:3setGeneralS}. Overall, the running time of the algorithm is $O(m n \min\{m,n\} + rk^2n)$.

\subsection{ Proof of Theorem \ref{thm3}}
We state the corresponding algorithm as Algorithm~\ref{alg:feature3}.
To prove Theorem~\ref{thm3},
we
 start with the general bound of Lemma~\ref{lem:generic2}; to apply the lemma,
we need to satisfy the rank assumption, which will become clear shortly,
during the course of the proof.

The algorithm of Theorem~\ref{thm3} constructs the matrix $\matZ$ by using
the algorithm of Lemma~\ref{tropp2} with $\epsilon$ set to a constant,
$\epsilon = \frac12.$
Using the same notation as in Lemma~\ref{tropp2},
$$\matE = \matA - \matA \matZ \matZ\transp,$$
and
$$\Expect{\FNormS{\matE}} \leq {\textstyle\frac{3}{2}}
\FNormS{\matA - \matA_k}.$$
We can now apply the Markov's inequality, to obtain that with probability at least $0.9$,
\begin{equation}\label{eqn:1}
\FNormS{\matE} \leq 15 \FNormS{\matA - \matA_k}.
\end{equation}

\begin{algorithm}[t] \label{alg:feature3}
\begin{framed}
\caption{{\sf Randomized Unsupervised Feature Selection}  (Theorem~\ref{thm3}) }
{\bf Input:} \math{\matA \in \R^{m \times k}}, number of clusters $k$,
and number of features $k < r < 4 k \log k$. \\
{\bf Output:} \math{\matC \in \R^{m \times r}} containing $r$ rescaled columns of $\matA$.
\begin{algorithmic}[1]
\STATE Compute the matrix \math{\matZ \in \R^{n \times k}} from the approximate SVD of $\matA$
in Lemma~\ref{tropp2}: $\matZ = FastApproximateSVD(\matA, k, \frac{1}{2})$ .
\STATE Let $c = \max\bigl\{r,16 k \log (20 k)\bigr\}$.
\STATE Let $[\matOmega_1,\matS_1]={\sf RandomizedSampling}(\matZ\transp, c).$ (see Algorithm~\ref{alg:sample3})
\STATE Compute $\tilde{\matV} \in \R^{c \times k}$, the matrix of
top $k$ right singular
vectors of $\matZ\transp\matOmega_1\matS_1 \in \R^{k \times c}$.
\STATE Let $[\matOmega,\matS] ={\sf DeterministicSamplingII}(\tilde{\matV}\transp, \matI_{c}, r). $ (see Algorithm~\ref{alg:sample2})
\STATE {\bf return} \math{\matC=\matA\matOmega_1\matS_1\matOmega\matS \in \R^{m \times r}.}
\end{algorithmic}
\end{framed}
\end{algorithm}

The randomized construction in the third step of Algorithm~\ref{alg:feature3}
gives sampling and rescaling matrices $\matOmega_1$ and $\matS_1$; the deterministic
construction in the fifth step of Algorithm~\ref{alg:feature3}
gives sampling and rescaling matrices $\matOmega$ and $\matS$.
To apply Lemma~\ref{lem:generic2}, we will choose
$$\matX_{in} = \matX_{opt},$$
since the lemma gives us the luxury to pick any indicator matrix $\matX_{in}$.
Note that we do not need to actually compute
\math{\matX_{opt}} in the algorithm; we are just using it in the proof
to get the desired result, as in the proof of Theorem~\ref{thm2}.

Algorithm~\ref{alg:feature3} first selects
\math{c} columns using $\matOmega_1\matS_1 \in \R^{n \times c}.$
Let $\matY = \matZ\transp \matOmega_1 \matS_1 \in \R^{k \times c},$
and consider its
SVD,
$$\matY = \tilde{\matU} \tilde{\matSig} \tilde{\matV}\transp,$$
with $\tilde{\matU} \in \R^{k \times k}$, $\tilde{\matSig} \in \R^{k \times k}$,
and $\tilde{\matV} \in \R^{c \times k}$.
Lemma~\ref{theorem:3setGeneralFr} now implies that with probability $0.9$,
\mld{
\sigma_k^2(\matY) \ge 1-\sqrt{\frac{4k\log(20k)}{c}}=\frac12,\label{eq:thm3-0}
}
because $c=16 k\log(20k).$
This means that
$\rank(\tilde\matV\transp)=k.$
Since \math{\tilde\matV\transp} is the input to
{\sf Deterministic Sampling \RN{2}}, and because \math{r>k},
it follows from
Lemma~\ref{theorem:3setGeneralS} that
\mld{\sigma_k(\tilde\matV\transp\matOmega\matS)>1-\sqrt{k/r}>0.\label{eq:thm3-00}}
Hence,
$$\sigma_k(Z\transp\matOmega_1\matS_1\matOmega\matS)=\rank(\matY\matOmega\matS)=\rank(\tilde\matV\transp\matOmega\matS)=k,$$
and we can apply Lemma~\ref{lem:generic2},
with  \math{\matE=\matA-\matA\matZ\matZ\transp}:
\mld{
\FNormS{ \matA -  \matX_{out} \matX_{out}\transp \matA} \le \FNormS{\matE} + 2 \gamma \frac{ \norm{(\matA -\matX_{opt} \matX_{opt}\transp \matA)
\matOmega_1 \matS_1 \matOmega \matS}_\mathrm{F}^2   +
\norm{\matE \matOmega_1 \matS_1 \matOmega \matS}_\mathrm{F}^2}
{ \sigma_k^2(\matZ\transp\matOmega_1\matS_1\matOmega \matS) }.
\label{eq:thm3-1}
}
To
bound the denominator of the second term, observe that
$$\sigma_k^2(\matZ\transp\matOmega_1\matS_1\matOmega \matS)=\sigma_k^2(\tilde{\matU} \tilde{\matSig} \tilde{\matV}\transp \matOmega \matS)
=
\sigma_k^2( \tilde{\matSig} \tilde{\matV}\transp \matOmega \matS),$$
where in the last equality we dropped \math{\tilde\matU}, which is
allowed since it is a full rotation.
Now, we obtain
\mld{
\sigma_k^2(\matZ\transp\matOmega_1\matS_1\matOmega \matS)=
\sigma_k^2( \tilde{\matSig} \tilde{\matV}\transp \matOmega \matS)
\ge \sigma_k^2(\tilde{\matSig}) \sigma_k^2( \tilde{\matV}\transp \matOmega \matS)
=
\sigma_k^2(\matY)
\sigma_k^2( \tilde{\matV}\transp \matOmega \matS).
\label{eq:thm3-2}
}
The last equality is because the singular values of
\math{\tilde\matSig} are exactly the singular values of
\math{\matY=\tilde\matU\tilde\matSig\tilde\matV\transp}. The first inequality
follows from the fact that for any two matrices $\matA$ and $\matB$ where $\matA$ has full column rank and $\matB$ has full row rank:
$\sigma_{\min}(\matA \matB) \ge \sigma_{\min}(\matA) \sigma_{\min}(\matB )$.
To prove this fact, notice
that $\sigma_{\min}(\matA) = \frac{1}{\TNorm{\matA^+}}$,
$\sigma_{\min}(\matB) = \frac{1}{\TNorm{\matB^+}}$, and
$\sigma_{\min}(\matA \matB) = \frac{1}{\TNorm{(\matA \matB)^+}} = \frac{1}{\TNorm{\matB^+ \matA^+}}$, where
the latter equality follows because if $\matA$ is full column rank and $\matB$ full row rank, then $(\matA \matB)^+ = \matB^+ \matA^+$~\cite{Gre66}.
By submultiplicativity $\TNorm{\matB^+ \matA^+} \le \TNorm{\matB^+} \TNorm{\matA^+}$. In our case,
$\matA=\tilde{\matSig}$ and $\matB = \tilde{\matV}\transp \matOmega \matS$.

Thus, using \r{eq:thm3-0} and \r{eq:thm3-00} in \r{eq:thm3-2}, we get
that with probability at least \math{0.9},
\mld{\sigma_k^2(\matZ\transp\matOmega_1\matS_1\matOmega \matS)\ge
{\textstyle\frac12}(1 - \sqrt{k/r})^2.\label{eq:thm3-2-1}}
We now bound
\math{\norm{
(\matA -\matX_{opt}
 \matX_{opt}\transp \matA) \matOmega_1 \matS_1 \matOmega \matS}_\mathrm{F}^2}:
\eqar{
\norm{(\matA -\matX_{opt} \matX_{opt}\transp \matA)
\matOmega_1 \matS_1 \matOmega \matS}_2^2
&=&
\norm{(\matA -\matX_{opt} \matX_{opt}\transp \matA)
\matOmega_1 \matS_1 \matI_c\matOmega \matS}_2^2 \nonumber\\
&\le&
\norm{(\matA -\matX_{opt} \matX_{opt}\transp \matA)
\matOmega_1 \matS_1}_\mathrm{F}^2\cdot\norm{\matI_c\matOmega \matS}_2^2 \nonumber\\
&\le&
(1+\sqrt{c/r})^2\norm{(\matA -\matX_{opt} \matX_{opt}\transp \matA)
\matOmega_1 \matS_1}_\mathrm{F}^2,\label{eq:thm3-3}
}
where the first inequality is by spectral submultiplicativity and
the second is because \math{\matI_c} is the input to
{\sf DetrministicSampling~\RN{2}} with \math{\ell_2=c}, and so
\math{\norm{\matI_c\matOmega \matS}_2^2\le1+\sqrt{c/r}}.
Similarly, we bound \math{\norm{
\matE \matOmega_1 \matS_1 \matOmega \matS}_\mathrm{F}^2}:
$$ \norm{
\matE \matOmega_1 \matS_1 \matOmega \matS}_\mathrm{F}^2 \le
(1+\sqrt{c/r})^2\norm{\matE
\matOmega_1 \matS_1}_\mathrm{F}^2.$$
From Lemma~\ref{theorem:3setGeneralFr},
$$
\Expect{ \FNormS{\matE \matOmega_1 \matS_1} } =
 \FNormS{\matE},$$
and
$$\Expect{ \norm{(\matA -\matX_{opt} \matX_{opt}\transp \matA)\matOmega_1 \matS_1}_\mathrm{F}^2}
=
\norm{\matA -\matX_{opt} \matX_{opt}\transp \matA}_\mathrm{F}^2.
$$
By a simple application of Markov's inequality and the union bound,
both of the equations below hold with probability at least $0.6$,
\eqar{
\FNormS{\matE \matOmega_1 \matS_1} &\le&
 5 \FNormS{\matE};\label{eq:thm3-4}\\
\norm{(\matA -\matX_{opt} \matX_{opt}\transp \matA)\matOmega_1 \matS_1}_\mathrm{F}^2
&\le& 5\norm{\matA -\matX_{opt} \matX_{opt}\transp \matA}_\mathrm{F}^2.
\label{eq:thm3-5}
}
Using \r{eq:thm3-2-1},\r{eq:thm3-3},\r{eq:thm3-4} and \r{eq:thm3-5}
in \r{eq:thm3-1}, together with a union bound, we obtain that with probability
at least \math{0.5}
\mld{
\FNormS{ \matA -  \matX_{out} \matX_{out}\transp \matA} \le \FNormS{\matE} + 20 \gamma \left(\frac{1+\sqrt{c/r}}{1-\sqrt{k/r}}\right)^2
\left(\norm{\matA -\matX_{opt} \matX_{opt}\transp \matA}_\mathrm{F}^2   +
\norm{\matE}_\mathrm{F}^2\right).
\label{eq:thm3-1}
}
We now use~\r{eqn:1} in \r{eq:thm3-1} and the fact that
\math{\FNormS{\matA-\matA_k}\le
\norm{\matA -\matX_{opt} \matX_{opt}\transp \matA}_\mathrm{F}^2};
since~\r{eqn:1} holds with probability at least 0.9 and
\r{eq:thm3-1} holds with probability at least 0.5, using a union bound,
we conclude that with probability at least
\math{0.4},
\mand{
\FNormS{ \matA -  \matX_{out} \matX_{out}\transp \matA} \le
\left(15+ 320 \gamma \left(\frac{1+\sqrt{c/r}}{1-\sqrt{k/r}}\right)^2\right)
\norm{\matA -\matX_{opt} \matX_{opt}\transp \matA}_\mathrm{F}^2.
}
The result follows because
\math{\cl F_{opt}=\norm{\matA -\matX_{opt} \matX_{opt}\transp \matA}_\mathrm{F}^2}
and \math{c=16k\log(20 k)}.
(Note, we have made no attempt to optimize the constants.)

The running time of Algorithm~\ref{alg:feature3} is $O\left(m n k + r k^3 \log(k) + r \log r \right)$,
since it employs the approximate SVD of Lemma~\ref{tropp2}, the randomized technique of Lemma~\ref{theorem:3setGeneralFr},
and the deterministic technique of Lemma~\ref{theorem:3setGeneralF}.

\section{Related work} \label{sxn:priorwork}

Feature selection has received considerable attention in the
machine learning and pattern recognition communities. A large number of
different techniques appeared in prior work, addressing
feature selection within the context of both clustering and
classification. Surveys include~\cite{GE03}, as well
as~\cite{GGBD05}, which reports the results of the NIPS 2003
challenge in feature selection. Popular feature selection
techniques include the Laplacian scores \cite{HCN06}, the Fisher
scores \cite{FS75}, or the constraint scores \cite{ZCZ08}.
None of these feature selection algorithms
have theoretical guarantees
on the performance of the clusters obtained using the
dimension-reduced
features.

We focus our discussion of related work on the family of feature selection
methods that resemble our approach, in that
they  select features by looking at the right singular vectors of
the data matrix \math{\matA}.
Given the input  $m \times n$ object-feature matrix $\matA$, and a
positive integer $k$,
a line of research tries to construct
features for (unsupervised) data
reconstruction, specifically for Principal Components
Analysis (PCA). PCA corresponds to the task of identifying a subset of
$k$ linear combinations of columns from $\matA$
that best reconstruct \math{\matA}. Subset selection for PCA asks to
find the columns of $\matA$ that reconstruct $\matA$ with comparable error as
do its top Principal Components.
Jolliffe~\cite{Jol72}
surveys various methods for the above task. Four of them (called
$B1$, $B2$, $B3$, and $B4$ in~\cite{Jol72}) employ the Singular
Value Decomposition of $\matA$ in order to identify columns that are
somehow correlated with its top $k$ left singular vectors. In
particular, $B3$ employs a deterministic algorithm which is very
similar to Algorithm~\ref{alg:feature3} that we used in this work;
no theoretical results are reported. An experimental
evaluation of the methods of~\cite{Jol72} on real datasets
appeared in~\cite{Jol73}. Another approach employing the matrix of
the top $k$ right singular vectors of $\matA$ and a Procrustes-type
criterion appeared in~\cite{Krz87}. From an applications
perspective,~\cite{WG05} employed the methods of~\cite{Jol72}
and~\cite{Krz87} for gene selection in microarray data analysis.

Feature selection for clustering
seeks to identify those features that have the most discriminative
power among all the features.~\cite{LSZT07} describes a method where one
first computes the matrix $\matV_k \in
\R^{n \times k}$, and then clusters the rows of $\matV_k$ by running, for
example, the $k$-means algorithm. One finally selects those $k$
rows of $\matV_k$ that are \emph{closest} to the centroids of the clusters
computed by the
previous step. The method returns those columns from $\matA$ that
correspond to the selected rows from $\matV_k$. A different approach
is described in~\cite{CD}. The method in~\cite{CD} selects features one at a
time; it first selects the column of $\matA$ which is \emph{most
correlated} with the top left singular vector of $\matA$, then
projects $\matA$ to this singular vector, removes the projection from
$\matA$, computes the top left singular vector of the resulting
matrix, and selects the column of $\matA$ which is \emph{most
correlated} with the latter singular vector, etc. Greedy
approaches similar to the method of~\cite{CD} are described in
\cite{MG04} and \cite{Mao05}. There are no known theoretical guarantees
for any of these methods. While these methods
 are superficially similar to our method,
in that they use the right singular matrix \math{\matV_k} and are
based on some sort of greedy algorithm, the techniques
we developed to obtain theoretical guarantees are entirely different and
based on linear-algebraic sparsification results~\cite{BSS09,BDM11a}.

The result most closely related to ours is the work in \cite{BMD09c,BZMD11}.
This work provides a randomized algorithm which offers a theoretical guarantee.
Specifically, for \math{r=\Omega(k\epsilon^{-2}\log k)}, it is possible
to select \math{r} features such that the
optimal clustering in the reduced-dimension space is a
\math{(3+\epsilon)}-approximation to the optimal clustering. Our result
improves upon this in two ways. First, our algorithms are deterministic; second, by using our
deterministic algorithms in combination with this randomized algorithm, we
can select \math{r=O(k)} features and obtain a competitive theoretical
guarantee.

Next, we should mention that if one allows linear combinations of the
features (feature \emph{extraction} rather than
feature \emph{selection}), then there are algorithms that offer
theoretical guarantees. First there is the SVD itself, which constructs
\math{k} (mixed) features for which the optimal clustering in this feature
space is a \math{2}-approximation to the optimal clustering~\cite{DFKVV99}.
It is possible to improve the efficiency of this SVD algorithm considerably
by using  the approximate SVD (as in Lemma~\ref{tropp2})
instead of the exact SVD to get nearly the
same approximation guarantee with \math{k} features. The exact statement
of this improvement can be found in~\cite{BZMD11}.
Boutsidis et al.~\cite{BZD10} show how to select \math{O(k\epsilon^{-2})}
(mixed) features with random projections and also obtaining a \math{(2+\epsilon)}-guarantee.
While these algorithms are interesting, they do not produce features that
preserve the integrity of the original features. The focus of this
work is on what one can achieve while preserving the original features.

A complementary line of research~\cite{PK05, HM04, FS06, FMC07, AV07, ADK09}
approaches the $k$-means problem by sub-sampling the points
of the dataset; such a subset of points is called \emph{coreset}.
The idea here is to select a small subset of
the points and by using only this subset obtain a partition
for all the points that is as good as the partition that would
have been obtained by using all the points.
\cite{PK05, HM04, FS06, FMC07, AV07, ADK09} offer algorithms for $(1+\epsilon)$
approximate partitions. Note that we were able to give only constant
factor approximations. For example, \cite{FS06} shows the existence
of an $(1+\epsilon)$-approximate coreset of size $r = O(k^3 / \epsilon^{n+1})$
($n$ is the number of features). \cite{FMC07} provides a coreset of size $r = poly(k, \epsilon^{-1})$.
The techniques used for all these coresets are different from the techniques we used
for feature selection; further, there is no clear way on how to use a coreset
selection algorithm to select features.
Moreover, the authors of~\cite{JKS12} use the coreset-based
algorithm from~\cite{AV07} to design a PTAS for k-means and other clustering problems.
It would be interesting to understand whether the techniques from
\cite{PK05, HM04, FS06, FMC07, AV07, ADK09, JKS12} are useful for feature selection as well. In particular,
it appears that there is potential to obtain relative error feature selection $k$-means algorithms
by using such approaches.

\section*{Acknowledgements}
Christos Boutsidis acknowledges the support from XDATA program of the Defense Advanced Research Projects Agency (DARPA), administered through Air Force Research
Laboratory contract FA8750-12-C-0323.

\bibliographystyle{abbrv}
\bibliography{RPI_BIB}

\end{spacing}

\end{document}